\documentclass[conference]{IEEEtran}
\IEEEoverridecommandlockouts

\usepackage{booktabs}
\usepackage{enumitem}
\usepackage{makecell}
\usepackage{stfloats}
\usepackage{xspace}
\usepackage{xurl}
\newcommand{\sys}{MLKV\xspace}
\newcommand{\kvs}{FASTER\xspace}
\usepackage{tikz}
\newcommand{\tmark}{\textcolor{teal}{\checkmark}}
\newcommand{\txmark}{\textcolor{teal}{\checkmark\kern-1.1ex\raisebox{.7ex}{\rotatebox[origin=c]{125}{--}}}}
\newcommand*\circled[1]{\tikz[baseline=(char.base)]{
            \node[shape=circle,fill=.,inner sep=0pt] (char) {\color{-.}\textsf\footnotesize #1};}}
\usepackage{cite}
\usepackage{amsmath,amssymb,amsfonts}
\usepackage{algorithmic}
\usepackage{graphicx}
\usepackage{textcomp}
\usepackage{xcolor}
\def\BibTeX{{\rm B\kern-.05em{\sc i\kern-.025em b}\kern-.08em
    T\kern-.1667em\lower.7ex\hbox{E}\kern-.125emX}}
\begin{document}
\title{\sys: Efficiently Scaling up Large Embedding Model Training with Disk-based Key-Value Storage}

\author{
\IEEEauthorblockN{Yongjun He}
\IEEEauthorblockA{\textit{ETH Z\"{u}rich} \\
yongjun.he@inf.ethz.ch}
\and
\IEEEauthorblockN{Roger Waleffe}
\IEEEauthorblockA{\textit{UW-Madison} \\
waleffe@wisc.edu}
\and
\IEEEauthorblockN{Zhichao Han}
\IEEEauthorblockA{\textit{eBay} \\
zhihan@ebay.com}
\and
\IEEEauthorblockN{Johnu George}
\IEEEauthorblockA{\textit{Nutanix} \\
johnu.george@nutanix.com}
\and
\IEEEauthorblockN{Binhang Yuan}
\IEEEauthorblockA{\textit{HKUST} \\
biyuan@ust.hk}
\and
\IEEEauthorblockN{Zitao Zhang}
\IEEEauthorblockA{\textit{eBay} \\
zitzhang@ebay.com}
\and
\IEEEauthorblockN{Yinan Shan}
\IEEEauthorblockA{\textit{eBay} \\
yshan@ebay.com}
\and
\IEEEauthorblockN{Yang Zhao}
\IEEEauthorblockA{\textit{eBay} \\
yzhao5@ebay.com}
\and
\IEEEauthorblockN{Debojyoti Dutta}
\IEEEauthorblockA{\textit{Nutanix} \\
debojyoti.dutta@nutanix.com}
\and
\IEEEauthorblockN{Theodoros Rekatsinas*\thanks{* Currently at Axelera AI.}
}
\IEEEauthorblockA{\textit{ETH Z\"{u}rich} \\
trekatsinas@inf.ethz.ch}
\and
\IEEEauthorblockN{Ce Zhang}
\IEEEauthorblockA{\textit{University of Chicago} \\
cez@uchicago.edu}
}

\maketitle
\begin{abstract}
Many modern machine learning (ML) methods rely on embedding models to learn vector representations (embeddings) for a set of entities (embedding tables).
As increasingly diverse ML applications utilize embedding models and embedding tables continue to grow in size and number, there has been a surge in the ad-hoc development of specialized frameworks targeted to train large embedding models for specific tasks.
Although the scalability issues that arise in different embedding model training tasks are similar, each of these frameworks independently reinvents and customizes storage components for specific tasks, leading to substantial duplicated engineering efforts in both development and deployment.

This paper presents \sys, an efficient, extensible, and reusable data storage framework designed to address the scalability challenges in embedding model training, specifically data stall and staleness.
\sys augments disk-based key-value storage by democratizing optimizations that were previously exclusive to individual specialized frameworks and provides easy-to-use interfaces for embedding model training tasks.
Extensive experiments on open-source workloads, as well as applications in eBay’s payment transaction risk detection and seller payment risk detection, show that \sys outperforms offloading strategies built on top of industrial-strength key-value stores by 1.6-12.6$\times$.
\sys is open-source at \url{https://github.com/llm-db/MLKV}.
\end{abstract}


\section{Introduction}
\label{sec:intro}
Embedding models have always been an important machine learning (ML) method for a variety of domains, including recommendations~\cite{DLRM-Survey, AmazonDLRM, DCN}, knowledge graphs (KGs)~\cite{KGE-Survey, DistMult, Complex}, and graph neural networks (GNNs)~\cite{GNN-Survey, GraphSage, GAT, Benchtemp}.
In these domains, it is common to convert sparse, high-dimensional data (e.g., attributes of entities and the interactions between entities) into continuous, low-dimensional (e.g., tens to hundreds) vector representations (embeddings).
The growing importance of these domains necessitates even larger embedding models to deliver improved predictive performance for a diverse array of tasks~\cite{ScalingLaws}, such as ranking and click-through rate (CTR) prediction for recommendation, node classification, and link prediction in KGs and GNNs.
For instance, both Meta~\cite{ZionEX} and Kuaishou~\cite{PERSIA} use embedding models exceeding 10 TB in scale for ranking and CTR prediction.

The exponential growth in embedding model size has led to the development of a flurry of specialized frameworks~\cite{AIBox, XDL, HugeCTR, PBG, Marius, MariusGNN, PERSIA, ZionEX, BagPipe, DGL-KE, PyG, DGL, Hetu} for training large embedding models.
Although earlier studies~\cite{PS, Petuum} observed that machine learning models can be represented as key-value abstractions, modern disk-based key-value stores are rarely used for large embedding model training due to scalability issues (details in Section~\ref{subsec:motivation}).
Consequently, they focus on building proprietary storage components for specific tasks or only support scale-out solutions.

To scale up embedding model training, mainstream frameworks usually start by re-implementing basic storage functionalities in their proprietary in-memory embedding management components.
By further integrating application logic to customize their storage components, they can exploit the characteristics of different tasks to minimize data stalls and staleness.
Unfortunately, from the perspective of developers, the maintenance, evolution, and optimization of these frameworks require significant duplication of effort and expense.
From the perspective of users, the deployment of existing frameworks is rather difficult, requiring the extensive rewriting of their applications due to the tight coupling of processing and storage components in these frameworks.
Furthermore, their applications require different rewrites for different tasks and frameworks, although the common goal is to scale embedding models.
Overall, these lead to two key questions:
\begin{itemize}[left=2pt]
    \item How can we design a data storage framework that provides extensible and reusable data management for embedding model training?
    \item Can we efficiently scale up large embedding model training with modern disk-based key-value stores?
\end{itemize}

To answer these questions, we propose \sys, a data storage framework that provides easy-to-use and non-intrusive interfaces to existing frameworks for managing embedding representations and utilizes bounded staleness consistency and look-ahead prefetching to accelerate the out-of-core training of large embedding models.
In summary, the key contributions of this work are as follows:
\begin{itemize}[left=2pt]
    \item We highlight the major issues in scaling up embedding model training: data stalls and staleness, and identify the mismatch between existing and desirable storage systems for embedding model training tasks.
    \item We present \sys, a data storage framework for flexibly scaling up embedding model training.
    \sys offers a set of interfaces to help ML task-specific frameworks utilize key-value storage for reusable, extensible, and efficient data management in embedding model training.
    \item We implement \sys based on \kvs~\cite{FASTER}.
    \sys democratizes optimizations previously exclusive to individual specialized frameworks, specifically bounded staleness consistency and look-ahead prefetching.
    \item We compare \sys against the SoTA on open-source workloads and real-world workloads from eBay (i.e., payment transaction risk detection and seller payment risk detection) to demonstrate its performance and scalability and validate the effectiveness of various optimizations.
\end{itemize}
\section{Preliminaries and related work}
\label{subsec:emb}

\begin{figure}[t]
\centering
\includegraphics[width=\columnwidth]{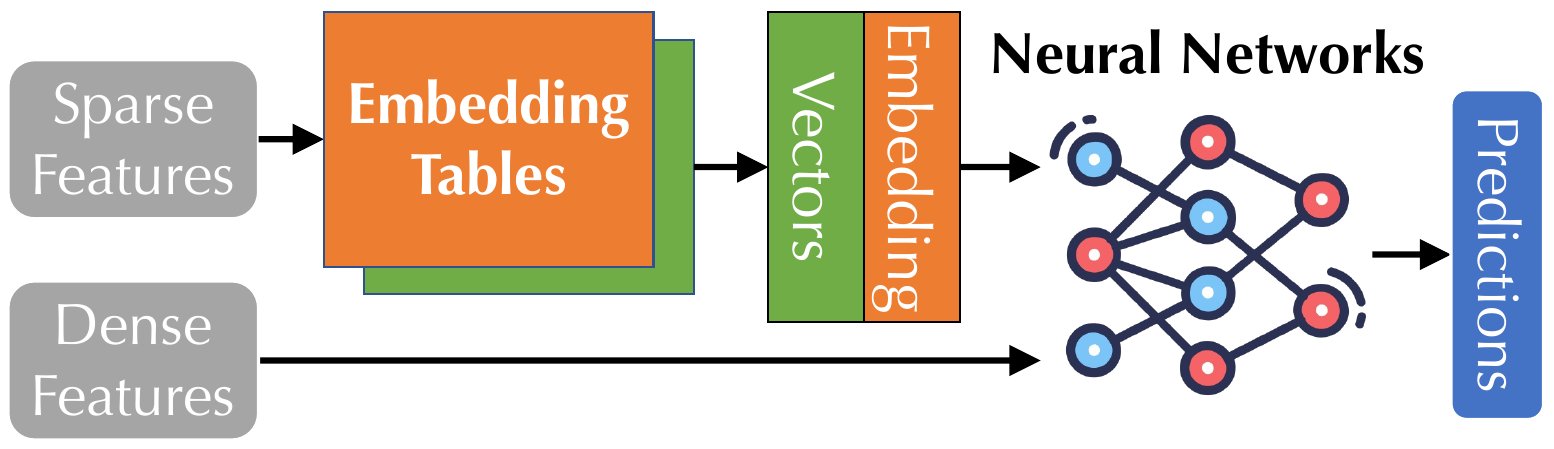}
\caption{Embedding table and neural network paradigm.}
\label{fig:paradigm}
\end{figure}

\subsection{Embedding Models}
To apply deep learning (DL) techniques to practical applications, one must adequately preprocess categorical data before feeding them into neural networks.
This makes the combination of embedding tables and neural networks a standard paradigm in various domains.
As shown in Figure~\ref{fig:paradigm}, sparse, high-dimensional features are first embedded into continuous, low-dimensional embedding vectors, and then fed into neural networks with dense features.

Formally, let $\mathcal{D} = \{ (x_\xi^S, x_\xi^D, y_\xi) \}_{\xi=1}^{|\mathcal{D}|}$ be the training dataset, where $\xi$ denotes the index of the training sample, $x_\xi^S := \{ x_{i;\xi}^S \mid x_{i;\xi}^S \in \mathbb{N}^{n_i} \}_{i=1}^m$ denotes the input sparse features from $m$ categorical fields, $n_i$ denotes the cardinality of the $i$-th category, $x_\xi^D$ denotes the input dense features, and $y_\xi$ denotes the label.
The corresponding embedding vectors of the $\xi$-th training sample can be computed by:
\begin{footnotesize}
\begin{align*}
E(w^{emb}; x_\xi^S) = (w_1^{emb}\, x_{1;\xi}^S, w_2^{emb}\, x_{2;\xi}^S, \dots, w_m^{emb}\, x_{m;\xi}^S)
\end{align*}
\end{footnotesize}
where the function $E(\cdot)$ denotes the embedding tables, $w_i^{emb} \in \mathbb{R}^{d \times n_i}$ denotes parameter of the $i$-th embedding table and $d$ denotes the embedding dimension.
The training objective of a given task can be formulated as:
\begin{footnotesize}
\begin{align*}
\min_{w^{emb}, w^{nn}} f(w^{emb}, w^{nn}) := \mathbb{E}_\xi \, \mathcal{L} ( N (w^{nn}; E(w^{emb}; x_\xi^S), x_\xi^D), y_\xi)
\end{align*}
\end{footnotesize}
where the function $N(\cdot)$ denotes the neural network, $w^{nn}$ denotes the parameter of the neural network, $\mathcal{L}$ denotes the loss function.

Large embedding models are usually trained asynchronously~\cite{AIBox, XDL, PERSIA, ZionEX, PBG, DGL-KE, Marius, PyG, DGL, MariusGNN, Hetu} to hide \textit{data stalls}, allowing the current iteration to progress without waiting for the embedding updates from earlier iterations.
Formally, let $t$ be the index of the current iteration and $k(t)$ denotes some early iterations, the update of $w^{emb}$ and $w^{nn}$ are denoted by
\begin{footnotesize}
\begin{align*}
    w^{emb}_{t+1} = w^{emb}_{t} - \gamma \nabla_{w^{emb}}{\mathcal{L}(w^{emb}_{k(t)}; w^{nn}_{t}; \xi_{t})}\\
    w^{nn}_{t+1} = w^{nn}_{t} - \gamma \nabla_{w^{nn}}{\mathcal{L}(w^{emb}_{k(t)}; w^{nn}_{t}; \xi_{t})}
\end{align*}
\end{footnotesize}
where $\gamma$ is the learning rate and $\nabla{\mathcal{L}(\cdot)}$ is the gradient of a function $\mathcal{L}$.
By allowing \textit{staleness}~\cite{Hogwild!, BSP, SSP}, described as $s = t - k(t)$, asynchronous training improves training throughput but may negatively affect embedding quality.

\subsection{Related Work}
\label{subsec:related}
\begin{figure}[t]
    \centering
    \includegraphics[width=\columnwidth]{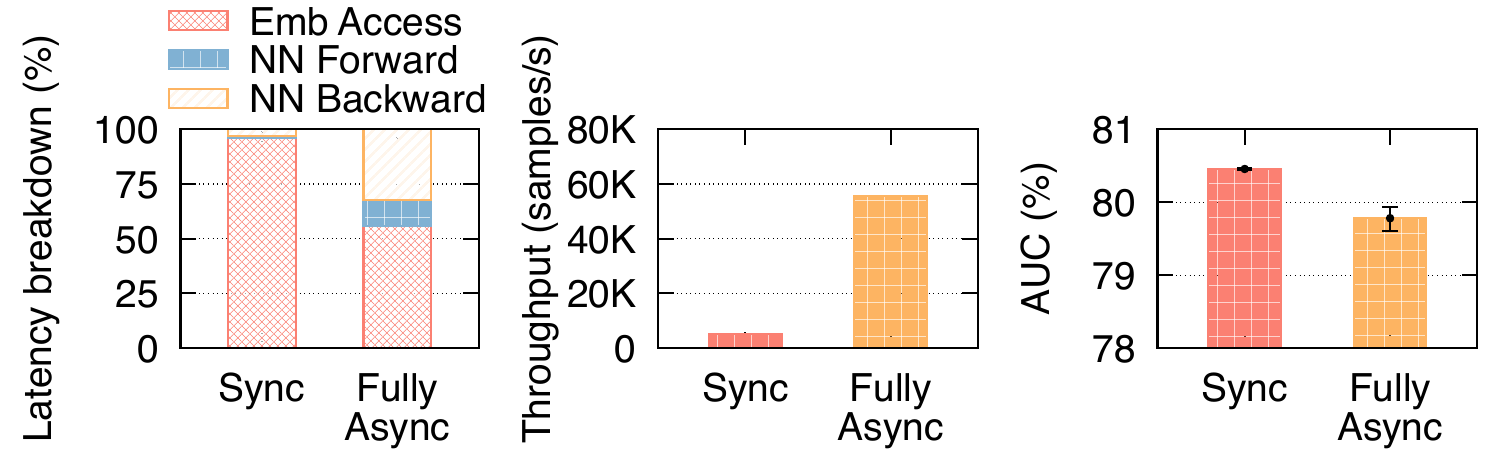}
    \caption{Scalability issues in embedding model training: (left and middle) poor throughput in synchronous training due to data stalls; (right) degraded model quality in fully asynchronous training due to staleness. We train DLRMs on the Criteo dataset using the PERSIA (as the computation layer) and FASTER (as the storage layer).}
    \label{fig:issues}
\end{figure}

\noindent \textbf{Embedding Model Training Frameworks.}
A flurry of frameworks has recently been developed to meet the emerging requirements of training embedding models larger than the GPU and CPU memory of a single machine.
PERSIA~\cite{PERSIA} scale up deep Learning recommendation models (DLRMs) to 100 trillion parameters by leveraging distributed memory management built on global hashing schemes and local LRU cache.
AIBox~\cite{AIBox} proposes a bi-level cache management system to accelerate the training of 10-TB scale DLRMs on a single node.
HugeCTR~\cite{HugeCTR} leverages RocksDB~\cite{RocksDB} for out-of-core DLRM inference.
PyG~\cite{PyG} allows direct stitching of graph learning tasks (e.g., GNNs and KGEs) with off-the-shelf key-value stores.
However, this approach leads to severe performance issues similar to those shown in Figure~\ref{fig:issues}.
Since the time we conducted our experiments and submitted our paper, DGL~\cite{DGL} has been concurrently developing a disk-based feature store for GNNs; moreover, it is still not compatible with DGL-KE~\cite{DGL-KE}.
PBG~\cite{PBG} and Marius~\cite{Marius, MariusGNN} divide knowledge graph embeddings (KGE) into multiple partitions and store each partition as a file.
Hetu~\cite{Hetu} is designed to support multiple tasks, but can not support larger-than-memory workloads.

\noindent \textbf{Parameter Server.}
This architecture~\cite{PS, Petuum, jiang-sigmod17, NuPS} uses a set of distributed in-memory key-value stores to provide consistency models and fault tolerance for distributed ML.
However, these scale-out solutions can not address the scalability issues in out-of-core training.
\sys focuses on scaling up large embedding model training with disk-based key-value stores, which has not been sufficiently explored in prior work.

\noindent \textbf{Disked-based Key-Value Storage.}
The simplicity and uniformity of the key-value interfaces (e.g., Get, Put, and Delete) make them easy to be reused as storage backbone by diverse systems~\cite{KVSE, cao2020, BlinkHash, Umbra, AthanassoulisIS23, CoroBase}.
B+tree~\cite{BDB, SQLite, WiredTiger, LMDB, LeanStore, TreeLine} based stores offer better read performance, while LSM-tree~\cite{FASTER, RocksDB, LevelDB, WiscKey} based stores offer better write performance.
\sys is based on \kvs~\cite{FASTER}, a log-structured store, but the optimizations we propose can also be applied to B+tree based key-value stores.

\noindent \textbf{Heterogeneous Storage.}
Recent studies show that the introduction of disk-based storage solutions in certain ML~\cite{AIBox, HugeCTR, ZeRO-infinity} and database~\cite{RocksMash} workloads can improve cost-effectiveness~\cite{simplyblock1, simplyblock2} compared to memory-centric or cloud-native storage solutions alone.
By periodically checkpointing to cloud-native storage, \sys can leverage the high performance of local NVMe SSDs while ensuring data persistence.

\begin{table}[t]
    \caption{Comparison of popular open-source frameworks. BS: Bounded staleness consistency. Ext: Extensibility. Reu: Reusability.}
    \begin{footnotesize}
    \begin{tabular}{p{0.15\linewidth}
                  p{0.08\linewidth}<{\centering}
                  p{0.05\linewidth}<{\centering}
                  p{0.05\linewidth}<{\centering}
                  p{0.09\linewidth}<{\centering}
                  p{0.06\linewidth}<{\centering}
                  p{0.02\linewidth}<{\centering}
                  p{0.03\linewidth}<{\centering}
                  p{0.03\linewidth}<{\centering}
                  }
    \toprule
    \textbf{Framework} &
    \textbf{DLRM} & \textbf{GNN} & \textbf{KGE} & \textbf{NoSQL} &
    \textbf{Disk} & \textbf{BS} &
    \textbf{Ext} & \textbf{Reu} \\
    \midrule
    PERSIA & \tmark & & & & & \tmark & & \\
    AIBox & \tmark & & & & \tmark & & & \\
    HugeCTR & \tmark & & & & \txmark & & & \\
    PyG & & \tmark & \tmark & & \txmark & & & \\
    PBG & & & \tmark & & \tmark & & & \\    
    DGL(-KE) & & \tmark & \tmark & & \txmark & & & \\
    Hetu & \tmark & \tmark & \tmark & & & \tmark & \tmark & \\
    \textbf{\sys} & \tmark & \tmark & \tmark & \tmark & \tmark & \tmark & \tmark & \tmark \\
    \bottomrule

    \end{tabular}
    \end{footnotesize}
    \label{tab:comparison}
\end{table}

\subsection{Problems and Opportunities}
\label{subsec:motivation}

\noindent \textbf{Problem 1:} Existing frameworks suffer from limited extensibility, poor reusability, and inadequate functionality (Table~\ref{tab:comparison}).
The tight coupling of storage management with application logic and the complexity of storage hierarchy implementation make these frameworks difficult to extend or reuse to meet new requirements and other tasks.
Though some of them consider extensibility~\cite{Hetu}, none of them can be reused without a massive rewrite of the applications to adopt their components.

\noindent \textbf{Problem 2:} Modern disk-based key-value stores suffer from scalability issues in large embedding model training (Figure~\ref{fig:issues}).
\begin{itemize}[left=2pt]
    \item Issue 1: Data stall. Synchronous training requires that the up-to-date embedding vectors are used to train the neural network, and the up-to-date gradients are used to update the embedding vectors.
    Therefore, the GPU remains idle during the fetching and updating of embedding vectors due to disk overheads, which is referred to as data stall.
    \item Issue 2: Staleness. Asynchronous training algorithms~\cite{Hogwild!, BSP, SSP} are thus gradually adopted to alleviate data stalls by overlapping data movement with computation.
    But it also introduces staleness to the embedding vectors, which may slow down the convergence rate and eventually decrease the quality of embedding models.
\end{itemize}

\noindent \textbf{Opportunities:} By addressing the above scalability issues, a data storage framework built on top of disk-based key-value storage can be efficient, extensible and reusable.
As revisited in Section~\ref{subsec:emb}, embedding tables are used to map sparse features into embeddings.
Hence, key-value interfaces can cleanly decouple the application logic from storage management: the computation layer uses the unique identifiers of sparse features to finish the application logic execution, and only invokes key-value interfaces to fetch and update the actual embeddings when neural networks require them.
\begin{figure}[t]
\begin{footnotesize}
\begin{verbatim}
1. import MLKV
2. ... # Application logic
3. nn_model, emb_tables = MLKV.Open(model_id, dim,
4.                                  staless_bound)
5. def train():
6.   for epoch in range(10):
7.     for (inputs, labels) in dataloader:
8.       ... # Application logic
9.       emb_values = emb_tables.Get(emb_keys)
10.       ... # Application logic
11.      output = nn_model(emb_values)
12.      loss = cross_entropy(output, labels)
13.      nn_optimizer.zero_grad()
14.      loss.backward()
15.      nn_optimizer.step()
16.      ... # Application logic
17.      emb_tables.Put(emb_keys, emb_values
18.        + emb_optimizer(emb_values.gradients))
19.      ... # Application logic
\end{verbatim}
\end{footnotesize}
\caption{Example usage of \sys.}
\label{fig:example}
\end{figure}

\section{\sys}
In this section, we describe the design and implementation of \sys.
\sys is a data storage framework that aims to: 1) support various mainstream embedding model training tasks; 2) support larger-than-memory workloads with competitive performance; 3) guarantee bounded-staleness consistency; and 4) retain extensibility and reusability, as shown in Table \ref{tab:comparison}.

\subsection{Interfaces Summary}
\sys offers users easy-to-use and non-intrusive interfaces to scale embedding model training tasks.
By simply creating, accessing, or modifying the embedding models with \sys interfaces in their applications, users can execute their embedding model training tasks as before.
An example usage of \sys is shown in Figure~\ref{fig:example}.
Next, we briefly describe the primary interfaces of \sys.

\begin{itemize}[left=2pt]
    \item The \textit{Open(model\_id, dim, staleness\_bound)} interface creates an embedding model with controllable \textit{staleness\_bound} (Section~\ref{subsec:opt}) and dimension.
    
    \item The \textit{Get(keys)} interface returns the \textit{values} (i.e., embedding vectors) associated with \textit{keys} (i.e., sparse feature identifiers) and is mainly used for forward propagation.

    \item The \textit{Put(keys, values)} upserts the \textit{values} associated with \textit{keys} and is mainly used for backward propagation.

    \item The \textit{Lookahead(keys, dest)} interface asynchronously loads the \textit{values} associated with \textit{keys} into the application cache or the memory buffer of \sys.
    Assisted with it, applications can leverage customized prefetching and caching strategies to hide and minimize disk accesses (Section~\ref{subsec:opt}).
\end{itemize}

\begin{figure}[t]
\centering
\includegraphics[width=0.8\columnwidth]{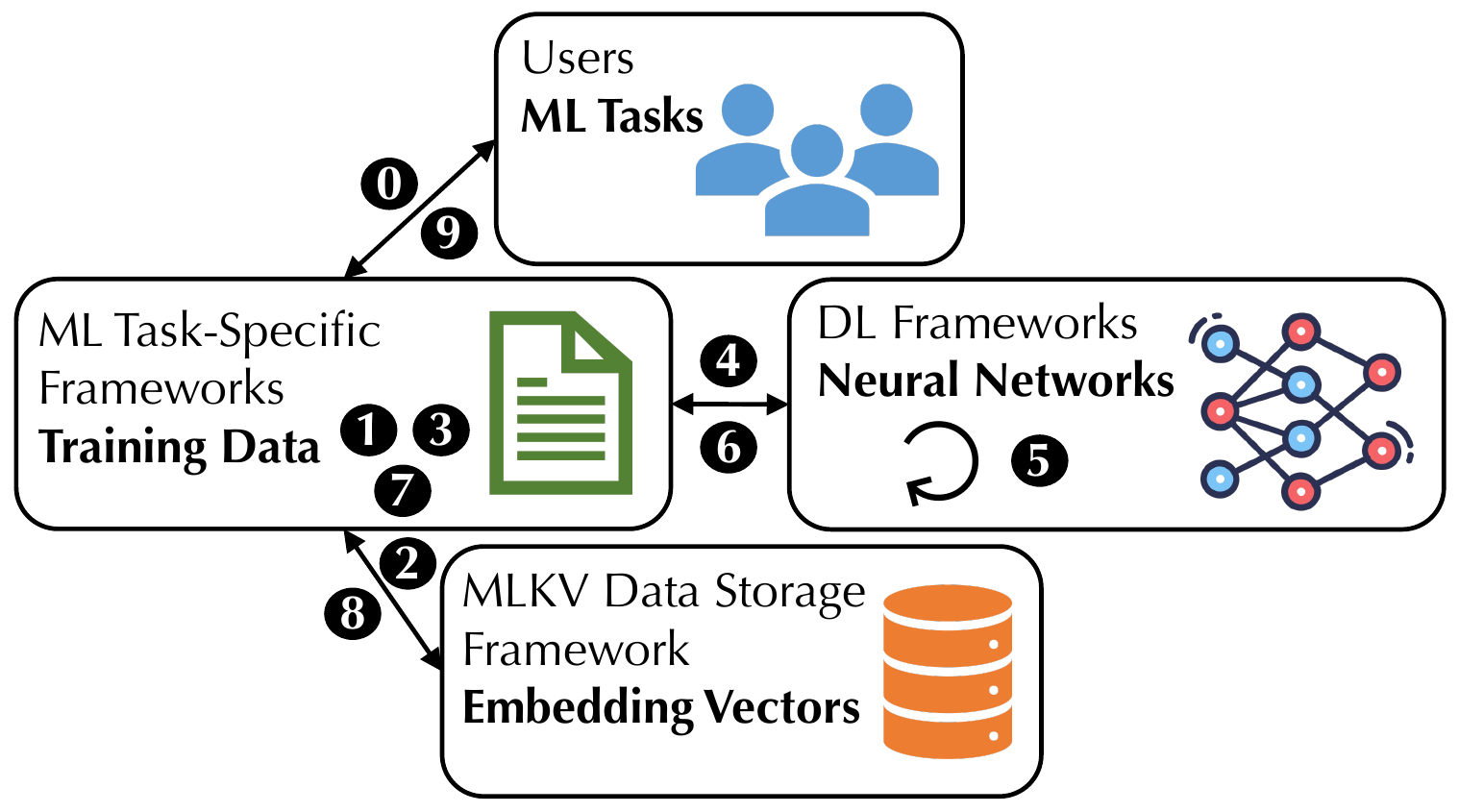}
\caption{Embedding model training with \sys.}
\label{fig:workflow}
\end{figure}

\subsection{Workflows}
We sketch the embedding model training workflow with \sys in Figure~\ref{fig:workflow}, where each step can be mapped to the code in Figure~\ref{fig:example}.
Once users start embedding model training tasks (step 
\circled{0}, line 2), ML task-specific frameworks then start to load training data and execute application logic (step \circled{1}, line 8)
to determine required sparse features.
After receiving requests containing identifiers for required sparse features, \sys invokes the Get interface to return corresponding embedding vectors (step \circled{2}, line 9).
ML task-specific frameworks then preprocess the embedding vectors according to the application logic (step \circled{3}, line 10)
before feeding them into the neural network (step \circled{4}, line 11).
The subsequent forward propagation, backpropagation, gradient synchronization, and model update are completely handled by DL frameworks (step \circled{5}, lines 12 - 15).
After receiving gradients of embedding tables (step \circled{6}, line 14), ML task-specific frameworks execute application logic to determine sparse features that will be updated (step \circled{7}, line 16).
After receiving requests containing identifiers and gradients for required sparse features, \sys invokes the Put or Rmw interface to update corresponding embedding vectors (step \circled{8}, line 17).
Users get results as before without extensively rewriting their applications (step \circled{9}, line 19).

\subsection{Implementation and Optimizations}
\label{subsec:opt}
Guided by a design goal of extensibility and reusability, \sys tweaks \kvs to democratize optimizations previously exclusive to individual custom-built frameworks.

\subsubsection{Bounded Staleness Consistency}
We augment \sys interfaces with user-configurable \textit{staleness\_bound} to address the staleness issue.
By configuring staleness\_bound to different values when creating embedding models, users can train their embedding model under different consistency models.
Embedding models are trained in Bulk Synchronous Parallel (BSP~\cite{BSP}) mode when staleness\_bound is set to 0 and are trained in Asynchronous Parallel (ASP~\cite{Hogwild!}) mode when staleness\_bound is set to infinity (in practice, INT64\_MAX).
For the rest configurations, embedding models are trained in Stale Synchronous Parallel (SSP~\cite{SSP}) mode.

\begin{figure}[t]
\centering
\includegraphics[width=1.0\columnwidth]{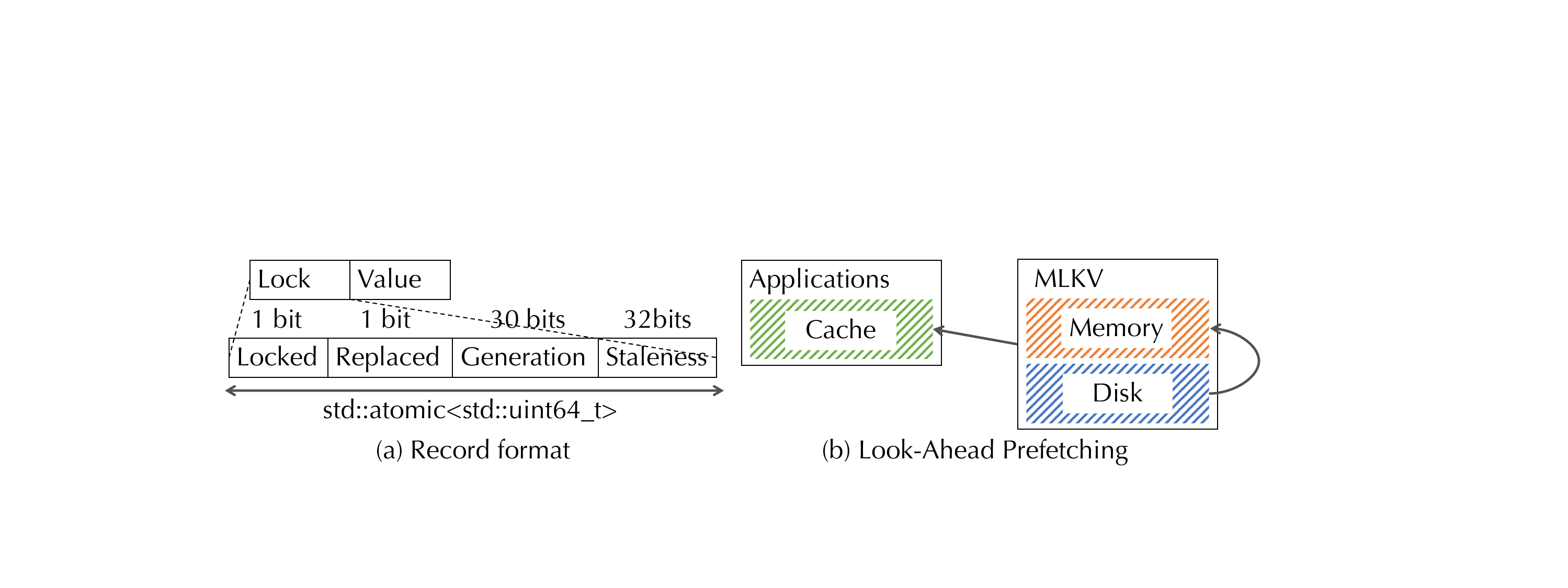}
\caption{Key designs of \sys.}
\label{fig:design}
\end{figure}

\sys provides bounded staleness consistency guarantees on a per-embedding basis, in other words, each key-value pair is associated with a vector clock.
Many latch-free key-value stores use 64-bit atomic variables as record-level locks to support concurrent access.
For example, \kvs uses 1 bit to indicate whether the record is locked, 1 bit to indicate whether the memory address of the record is replaced by other threads, and 30 bits to indicate the generation of the record to ensure that the latest value is returned.
\sys implements latch-free vector clocks on top of it by stealing unused bits in record-level locks to indicate staleness.
The detailed \sys record format is shown in Figure~\ref{fig:design}(a), and the implementation of Get and Put interfaces are introduced below.
Concurrent Get and Put operations first attempt to acquire the lock for the record associated with the key.
To acquire the lock, a Get operation must repeatedly check the staleness flag until it is less than the staleness bound, whereas a Put operation can skip this step because it only reduces the staleness.
Then they confirm the record is not locked, the memory address is not replaced, the generation is the latest, and swap the original lock to a desired lock, all of which can be done with a single atomic compare-and-swap.
The desired lock has the Locked bit set to 1 and increment or decrement the staleness flag by 1 for Get or Put operations, respectively.
Next, the Get operation will return the embedding vector for the forward propagation, and the Put operation will update the embedding vector for the backward propagation.
Finally, \sys will perform a fetch-and-sub to set the replaced bit and increment the generation flag by 1.

\subsubsection{Look-Ahead Prefetching}
Read-ahead is a technique used in database systems like MySQL~\cite{MySQL} and SQL Server~\cite{SQLServer} that predicts when pages will be needed and brings them into the buffer pool before they are actually used by the query.
Inspired by Read-Ahead, we propose look-ahead prefetching.
Unlike conventional prefetching, which retrieves embeddings within the staleness bounds into the application cache in advance, look-ahead prefetching further brings embeddings beyond the staleness bounds from the disk into the storage system's buffer pool ahead of time, thus avoiding the limitations imposed by staleness bounds.
Therefore, users can also use look-ahead prefetching to manipulate cache admissions for customized caching strategies.

For example, in a CTR task with a staleness bound of 4, even if the application knows the next 100 training samples, conventional prefetching can prepare embeddings for only up to 4 training samples in advance.
Now, with look-ahead prefetching, \sys can further hide the disk accesses of the remaining training samples while guaranteeing bounded staleness consistency.

We extend \sys with a non-blocking Lookahead interface that asynchronously loads embedding vectors from the disk to the application cache or its memory buffer (Figure~\ref{fig:design}(b)).
Once users or applications have customized caching strategies, or even just know what future incoming training samples will be, they can invoke the Lookahead interface to prefetch embedding vectors in advance for future use, hiding disk accesses during training.
When the destination is the application cache, look-ahead prefetching works the same as conventional prefetching, so below we only describe the process when the destination is \sys's memory buffer.
Once the value is obtained and there are no other threads updating it, a new record with the original staleness and value will be copied into the mutable memory buffer of \sys.
If the data is not on disk but in the immutable memory buffer, we will not copy it into the mutable memory, which can reduce the number of pages written to disk.
This is because \sys is built on log-structured merge-trees (LSM-trees) like data structure, where the current mutable memory buffer will be switched to immutable when it is full, and then written to disk.
\begin{table}[t]
  \caption{Datasets and models.}
  \centering
  \begin{footnotesize}
  \begin{tabular}{p{0.21\linewidth}
                  p{0.09\linewidth}
                  p{0.06\linewidth}
                  p{0.08\linewidth}
                  p{0.28\linewidth}
                  }
    \toprule
    \textbf{Dataset} & 
    \textbf{\# Emb} & \textbf{Dim} & 
    \textbf{Type} & \textbf{Model} \\
    \midrule

    Freebase86M & 
    86M & 100 & 
    KGE & DistMult \& ComplEx \\

    WikiKG2 & 
    2.5M & 400 & 
    KGE & DistMult \& ComplEx \\

    Papers100M & 
    111M & 128 & 
    GNN & GraphSage \& GAT \\

    eBay-Payout & 
    1.7B & 768 & 
    GNN & GraphSage \\

    eBay-Trisk & 
    185M & 256 & 
    GNN & GraphSage \\

    Criteo-Terabyte & 
    883M & 16 & 
    DLRM & FFNN \& DCN \\

    Criteo-Ad & 
    34M & 16 & 
    DLRM & FFNN \& DCN \\
    \bottomrule
  \end{tabular}
  \end{footnotesize}
  \label{tab:dataset}
\end{table}

\begin{figure*}[t]
    \centering
    \includegraphics[width=\textwidth]{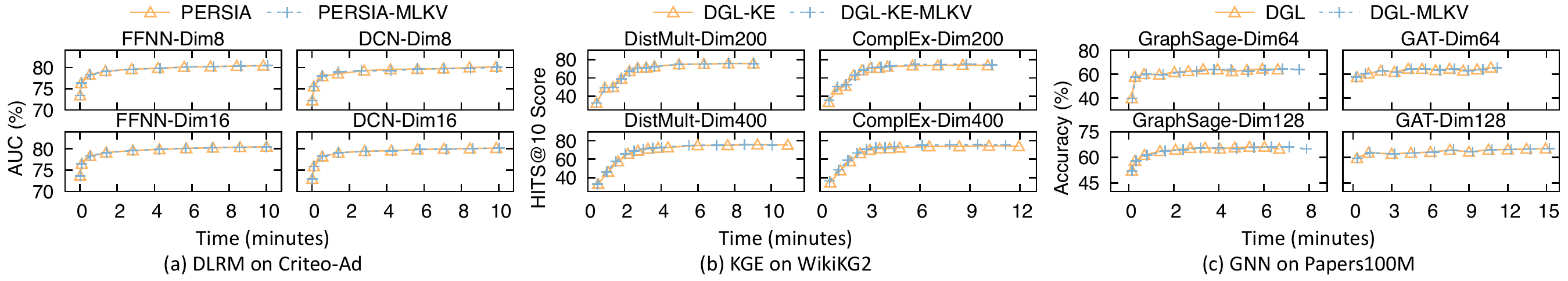}
     \caption{End-to-end training convergence comparison.}
    \label{fig:e2e}
\end{figure*}

\begin{figure*}[b]
    \centering
    \includegraphics[width=\textwidth]{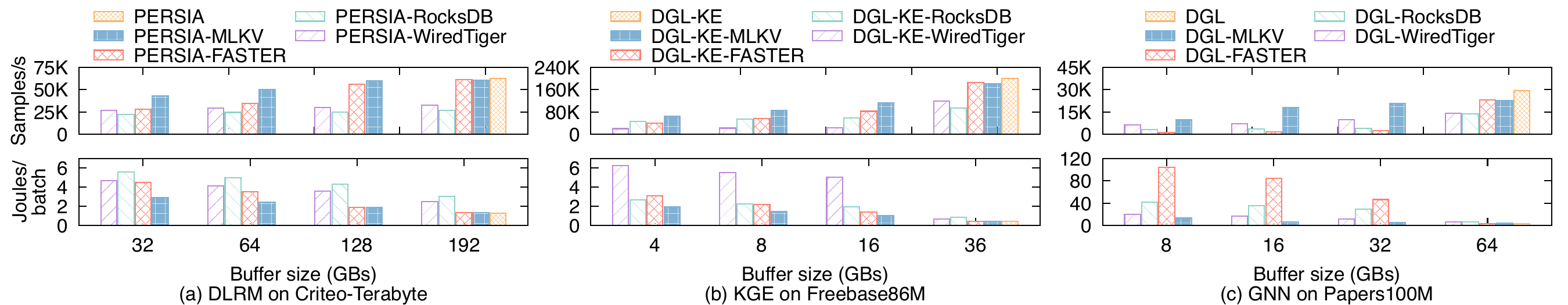}
    \caption{Impact of cache management on (top) throughput (higher is better) and (bottom) energy consumption (lower is better).}
    \label{fig:disk}
\end{figure*}

\section{Evaluation}
\label{sec:eval}

\subsection{Experimental Setup}
\noindent \textbf{Datasets.}
We evaluate the performance of \sys against baseline systems using seven datasets (Table~\ref{tab:dataset}), including five open-source datasets~\cite{Freebase, OGB, Criteo-AD, Criteo-Terabyte} and two real-world production datasets, eBay-Trisk and eBay-Payout.
eBay-Trisk is a payment transaction risk detection workload on a bipartite graph, which consists of 185 million nodes that represent either transactions or entities (e.g., buyers).
eBay-Payout is a seller payout risk detection workload on a graph consisting of 1.7 billion nodes representing sellers, items, or buyer checkouts.
 
\noindent \textbf{Hardware}.
Experiments using the open-source datasets were run on a single AWS g5.16xlarge instance, while experiments using the eBay datasets were run on eBay machines with one V100 GPU, 256 GB CPU memory, and SSDs with 1024 MB/s bandwidth per instance.

\noindent \textbf{Software.}
We compared \sys with six open-source systems: DGL~\cite{DGL} for GNN training, DGL-KE~\cite{DGL-KE} for KGE model training, PERSIA~\cite{PERSIA} for DLRM training, and the integration of the above three systems and FASTER~\cite{FASTER}, RocksDB~\cite{RocksDB}, and WiredTiger~\cite{WiredTiger} for larger-than-memory workloads.

\noindent \textbf{Tasks, Metrics, Models, and Hyperparameters.}
We use three representative embedding model training tasks to evaluate the end-to-end performance.
For DLRM training, we use the CTR task and models including fully connected feed-forward neural network (FFNN) and DCN~\cite{DCN},
and report Area under the ROC Curve (AUC).
For KGE model training, we use the link prediction task and models including DistMult~\cite{DistMult} and Complex~\cite{Complex},
and report the Hits@k score, which is the fraction of the top k entities that are ranked correctly.
For GNN training, we use the node classification task and models including GraphSage~\cite{GraphSage} and GAT~\cite{GAT}, and report accuracy and AUC.
We also evaluate \sys on YCSB~\cite{YCSB}, a NoSQL benchmark, to isolate the effects of application code.
For all the tasks, we also report throughput.

\subsection{End-to-End Comparison}
\label{subsec:e2e}
We first compare the convergence speed of \sys with the SoTA frameworks on three representative tasks.
In this experiment, we ensure that all variants share the same application logic and staleness bounds and that the embedding models fit in memory.
As shown in Fig~\ref{fig:e2e}, \sys can achieve the same convergence threshold as the corresponding baseline frameworks in a comparable time. 
The efficient support for various tasks demonstrates the extensibility and reusability of \sys.
Compared with the specialized frameworks with proprietary storage architectures, \sys is at most 2.5\%, 2.6\%, and 22.2\% slower than PERSIA, DGL-KE, and DGL due to the additional overhead of index traversal.

In the larger-than-memory workloads, we compare \sys with the integrations of the SoTA frameworks and industrial-strength key-value stores, following the same settings as above but using larger datasets and varying the buffer size.
Figure~\ref{fig:disk}(top) shows that the \sys-based solutions outperform other variants by 1.08-2.44$\times$ in the DLRM CTR task, 1.36-4.89$\times$ in the KGE link prediction task, and 1.53-12.57$\times$ in the GNN node classification task.
The \sys-based solutions can better hide data stalls while guaranteeing staleness bounds than the variants.
We also report approximate energy consumption following previous methods~\cite{roose2018predictions,Carbontracker,Zeus}.
Figure~\ref{fig:disk}(bottom) shows that the \sys-based solution is more energy-efficient than other solutions in larger-than-memory workloads.

\begin{figure}[t]
    \centering
    \includegraphics[width=\columnwidth]{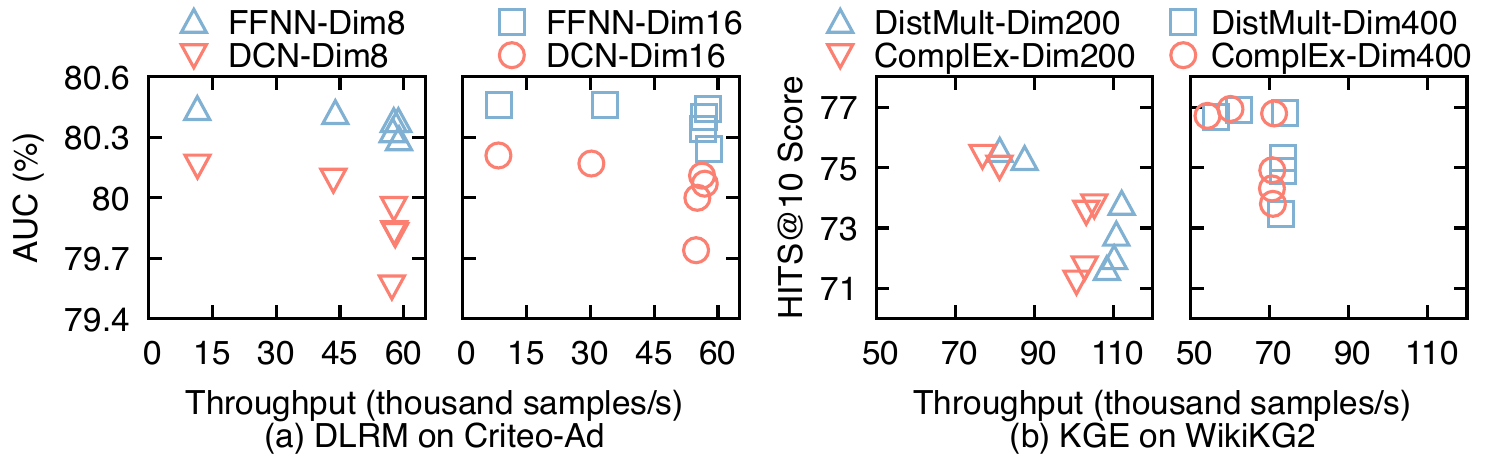}
    \caption{Effect of bounded staleness consistency.}
    \label{fig:staleness}
\end{figure}
\subsection{Effect of Bounded Staleness Consistency.}
\label{subsec:staleness}

To analyze the effect of bounded staleness consistency in isolation, we run the CTR task and the link prediction task with \sys by fixing the buffer size and varying the staleness bounds (0-80).
By appropriately relaxing the staleness bound, the \sys-based solutions can achieve up to 6.58$\times$ speedup with a tolerable degradation in model quality (i.e., less than 0.1\% drop of AUC), as demonstrated by the empirical results (Figure~\ref{fig:staleness}).
In contrast, the \kvs-based solutions inevitably cause the AUC to drop by more than 0.8\% (Figure~\ref{fig:issues}(right)).
A rigorous theoretical analysis of the convergence guarantees with regard to staleness bound is provided in~\cite{PERSIA}.

\begin{figure}[t]
    \centering
    \includegraphics[width=\columnwidth]{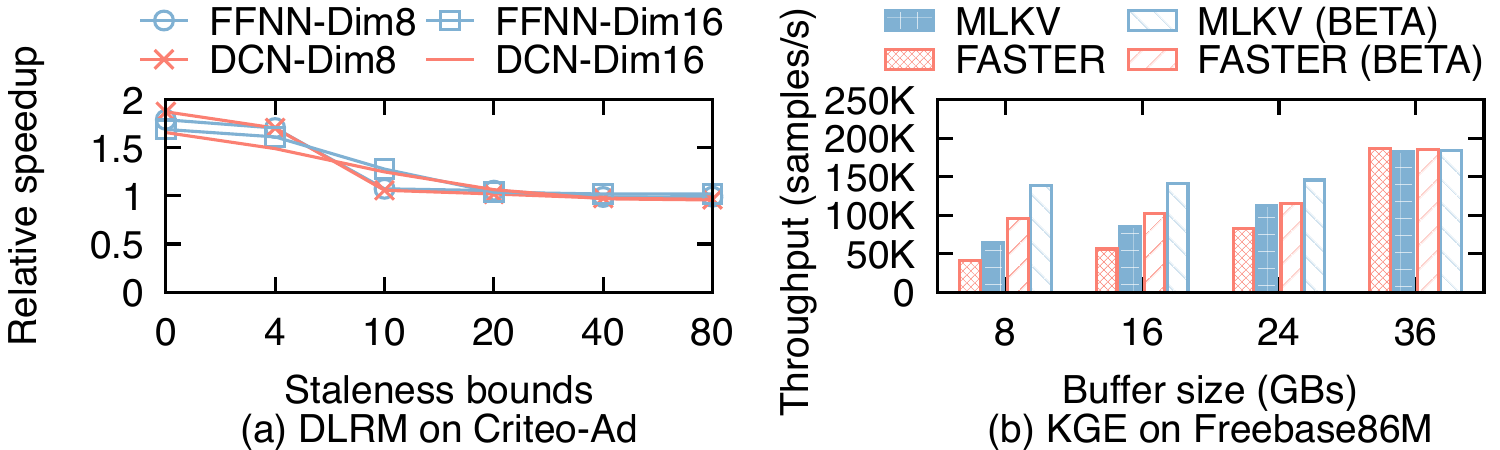}
    \caption{Effect of look-ahead prefetching.}
    \label{fig:lookahead}
\end{figure}

\subsection{Effect of Look-Ahead Prefetching.}
\label{subsec:lookahead}
The experiments follow the settings of Section~\ref{subsec:staleness}, but vary the buffer size and additionally evaluate a partition-based graphing learning algorithm, BETA~\cite{Marius,MariusGNN}, in the KGE model training.
When the staleness bounds are low, look-ahead prefetching can significantly improve the training throughput (Figure~\ref{fig:lookahead}(a)).
However, when the staleness bounds are high, conventional prefetching alone is enough to hide data stalls, so the effect of look-ahead prefetching is not that significant.
In the link prediction task, Figure~\ref{fig:lookahead}(b) shows that look-ahead prefetching can improve the training throughput of both standard and partition-based graph learning algorithms.

\label{subsec:nosql}
\begin{figure}[t]
    \centering
    \includegraphics[width=\columnwidth]{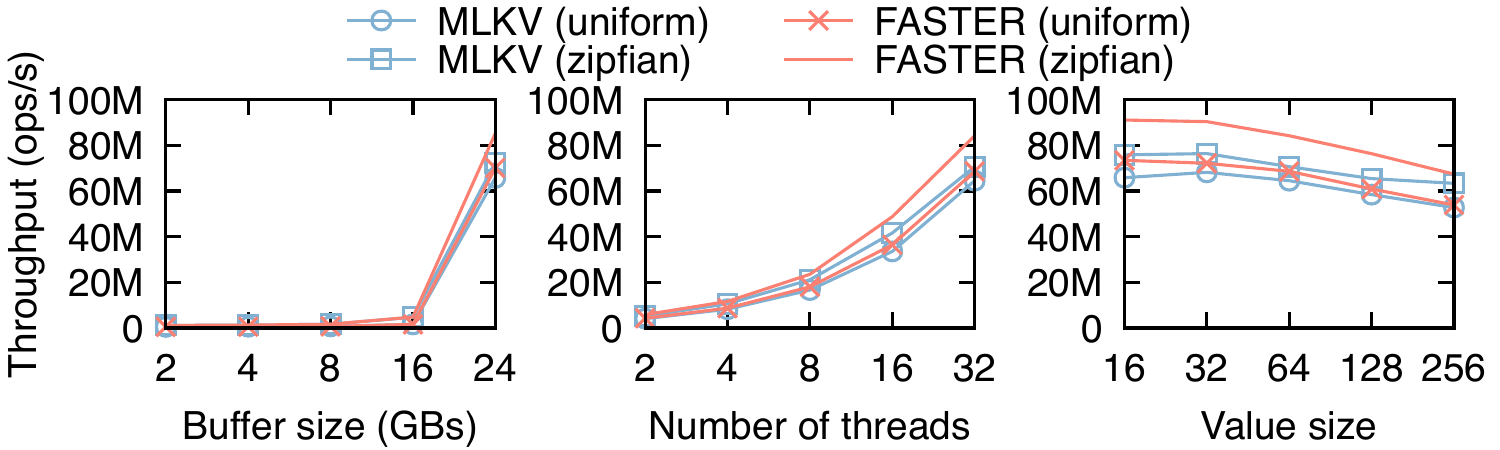}
    \caption{Throughput comparison between \sys and \kvs on YCSB workloads.
    }
    \label{fig:ycsb}
\end{figure}
\subsection{NoSQL Workloads}
To quantify the overhead brought by our implementation, we compare \sys with \kvs on YCSB workload with 50\% read operations and 50\% write operations.
For uniform access, the performance gap is less than 10\%, and for skewed access, the performance gap is less than 20\% (Figure~\ref{fig:ycsb}).
This is attributed to the fact that the overhead from the vector clock becomes more pronounced in skewed workloads.
If the user disables bounded stale consistency, \sys only incurs memory overhead and no performance overhead.

\begin{figure}[t]
    \centering
     \includegraphics[width=\columnwidth]{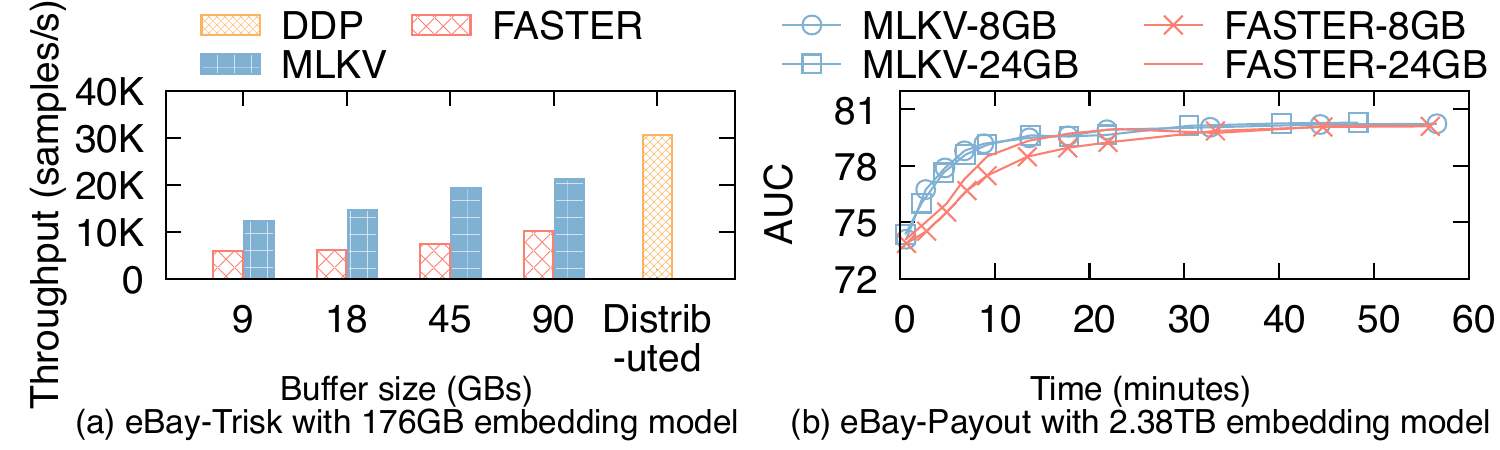}
    \caption{Application of \sys in risk detection at eBay.}
    \label{fig:ebay}
\end{figure}
\subsection{Case Studies: eBay-Trisk and eBay-Payout}
\label{subsec:ebay}

We now illustrate how \sys helps scale GNN training for the payment transaction risk detection
and seller payout risk detection at eBay.
The result reported in Figure~\ref{fig:ebay} leads to two findings:
1) Compared to DGL-DPP which requires two instances to hold the entire embedding model, DGL-\sys is more cost-effective since it achieves comparable training throughput (69.6\% of DGL-DDP) with only one instance.
2) Look-ahead prefetching effectively hides data stalls in larger-than-memory workloads, thereby improving training throughput.
\section{Conclusion}
Today's machine learning applications urgently require larger and more embedding models, resulting in many ad-hoc solutions with poor extensibility and reusability.
This paper presents \sys, a unified data storage framework that scales embedding model training with key-value storage.
\sys democratizes optimizations previously only implemented in individual custom-built frameworks while retaining extensibility and reusability.
Our evaluations show that \sys outperforms offloading strategies built on top of industrial-strength key-value stores by 1.6-12.6$\times$ for larger-than-memory workloads and closely matches the performance of specialized frameworks for in-memory workloads.
\section*{Acknowledgment}
We thank Gustavo Alonso for his invaluable comments.

\bibliographystyle{IEEEtran}
\bibliography{sample}

\end{document}